\documentclass[letterpaper]{article} 
\usepackage[preprint]{aaai2027}  
\usepackage[hyphens]{url}  
\usepackage{graphicx} 
\def\UrlFont{\rm}  
\usepackage{natbib}  
\usepackage{caption} 
\usepackage{algorithm}
\usepackage{algorithmic}
\usepackage{newfloat}
\usepackage{listings}
\DeclareCaptionStyle{ruled}{labelfont=normalfont,labelsep=colon,strut=off} 
\floatstyle{ruled}
\newfloat{listing}{tb}{lst}{}
\floatname{listing}{Listing}

\usepackage{booktabs}
\usepackage{multirow}
\usepackage{amsmath}
\usepackage{amssymb}
\usepackage[table]{xcolor}
\usepackage{xcolor}
\usepackage{soul}
\usepackage{subcaption}
\definecolor{softlinkblue}{RGB}{92,151,198}
\usepackage[
  colorlinks=true,
  linkcolor=softlinkblue,
  citecolor=black,
  urlcolor=black
]{hyperref}

\sethlcolor{yellow}
\definecolor{premrow}{RGB}{232,244,255}
\newcommand{\premcell}[1]{\cellcolor{premrow}#1}

\title{\textsc{PReM}: Learning What to Preserve and When to Refresh for Context Compression}
\author{
  Bohan Yu\thanks{\,\,\,Equal contribution.},
  Lei Shen\footnotemark[1]\thanks{\,\,\,Corresponding author.},
  Chenxi Zhou,
  Chen Han,
  Junlin Liu,\\
  Wenbo Su,
  Yu Cheng,
  Bo Zheng
}
\affiliations{
    Alibaba Group \\
    \texttt{yibo.bohanyu@gmail.com} \quad \texttt{shenl15@tsinghua.org.cn}
}

\begin{document}

\maketitle

\begin{abstract}
Efficient long-context inference is not only about reducing memory cost, but also about keeping useful contextual evidence accessible as generation proceeds.
However, existing compression-oriented approaches, such as key-value (KV) cache compression and context compression, often either make an early decision about which contextual information to keep or rely on an external compressor.
Such designs make it difficult to adapt the compressed context to the evidence needed by later reasoning steps.
This paper introduces PReM (\textbf{P}reserve and \textbf{Re}fresh \textbf{M}emory), a context-compression framework that maintains the long context as the model's internal layer-wise KV memory and learns what to preserve and when to refresh it.
Specifically, PReM uses a dedicated memory layer to make memory-selection decisions, and a special memory token \(\langle m\rangle\) to trigger refreshes during generation.
To train this behavior, PReM introduces \textit{Phase-Separated Refresh Training}, aligning memory selection with memory-conditioned generation while preserving continuity across refreshes.
Experiments with 32K-token contexts show that PReM outperforms strong baselines under both 16× and 32× compression, while maintaining a favorable balance between answer quality and inference efficiency.
\end{abstract}


\section{Introduction}

Long-context inference has become a natural way to use large language models (LLMs) for knowledge-intensive tasks: a model receives a large collection of contextual evidence and generates task-specific outputs from it~\cite{openai2024gpt4technicalreport,zhao2023survey}.
However, the inference cost grows quickly as the context expands~\cite{dao2022flashattentionfastmemoryefficientexact}.
Long inputs enlarge the key-value (KV) cache~\cite{pope2022efficientlyscalingtransformerinference} maintained by every attention layer, while each generated token must attend to this enlarged cache, substantially increasing decoding latency~\cite{kwon2023efficientmemorymanagementlarge}.
This creates a tension between efficiency and evidence access: reducing the cost of long-context inference should not make the model lose access to the contextual evidence needed during generation.

Existing approaches to efficient long-context inference, such as KV-cache compression and context compression, ease this tension only partially.
KV-cache methods~\cite{zhang2023h2oheavyhitteroracleefficient,ICLR2024_streamingllm,NEURIPS2024_snapkv,cai2025pyramidkvdynamickvcache,yu-chai-2025-evolkv,qin2025cake} reduce the cached states used during attention by retaining different subsets of KV states, often guided by positional, layer-wise, or attention-based properties.
Context-compression methods shorten the input before generation, either in text space~\cite{jiang-etal-2024-longllmlingua,pan-etal-2024-llmlingua,hwang2025exitcontextawareextractivecompression} or through learned soft memory tokens~\cite{ICLR2024_ICAE,ICLR2025_Activation_Beacon,tang2026comicoarsetofinecontextcompression}.
Despite their effectiveness, these methods still face two limitations.
\textbf{(1) Static evidence retention.} They typically make an early decision about which contextual information to retain and reuse the compressed representation throughout generation.
However, different reasoning steps, especially in multi-hop tasks, may depend on different contextual evidence, so this early decision may preserve irrelevant information while discarding evidence needed later.
\textbf{(2) Decoupled evidence selection.} Methods that rely on an external compressor~\cite{jiang-etal-2024-longllmlingua,hwang-etal-2025-exit,tang2026comicoarsetofinecontextcompression} perform compression separately from generation, introducing additional computation and making evidence selection less directly tailored to the needs of each reasoning step.
Taken together, these limitations suggest that effective compression should go beyond reducing context size: \textit{it should preserve useful contextual evidence and refresh it whenever later reasoning requires different evidence}.

To address this gap, we propose PReM (\textbf{P}reserve and \textbf{Re}fresh \textbf{M}emory), a context-compression framework for inference-time generation that learns what contextual evidence to preserve and when to refresh it.
The key idea is to maintain the long context as compressed memory that can be refreshed according to the model's current generation state and step-specific evidence needs.
Instead of relying on an external compressor, PReM instantiates this memory within the LLM itself, using cached layer-wise KV states as the memory representation.
This KV memory is partitioned into chunks, and PReM uses a dedicated \textit{memory layer} to make memory-selection decisions.
Specifically, selected chunks are preserved at token-level granularity, while unselected chunks are compressed into averaged KV representatives.
At inference time, the question provides the initial signal for memory selection before answer generation.
To make refresh for later reasoning steps controllable by the model itself, PReM further introduces a special memory token \(\langle m\rangle\) as an explicit internal refresh signal.
Whenever \(\langle m\rangle\) is emitted, the memory layer uses its hidden state to reselect chunks relevant to the current step, and this updated selection refreshes the compressed memory in other layers.
Because each refresh changes the KV memory seen by subsequent tokens, training must condition each generation step on the refreshed memory state.
PReM therefore uses \textit{Phase-Separated Refresh Training}, splitting each step into a \textit{memory-selection} phase and a \textit{memory-conditioned generation} phase with separate forward passes and losses.
The former is supervised to identify step-specific evidence chunks, whose selection refreshes the compressed memory, while the latter trains autoregressive generation under the resulting memory state.
An additional boundary-aware objective connects separated phases and adjacent steps.
Through this design, PReM dynamically preserves the evidence needed by each reasoning step and learns to refresh memory at appropriate moments.

Experiments with 32K-token contexts show that PReM substantially improves both answer quality and computational efficiency.
Under both 16× and 32× compression ratios, PReM outperforms eight strong methods spanning KV-cache and context compression on 3B and 7B backbones, improving average exact match and F1 by up to 10.23/12.55 points over the strongest baseline.
Its gains over full-context prompting indicate that learned memory selection can be more effective than exposing the base model to the entire 32K context.
Notably, the 3B PReM model even outperforms released 7B soft context-compression baselines, showing that learned preservation and refresh can compensate for smaller backbone scale.
Efficiency analyses further show that PReM reduces peak memory while maintaining competitive inference latency, demonstrating a favorable balance between performance and efficiency.

Our contributions are summarized as follows:
\begin{itemize}
    \item This paper formulates context compression as a learned memory preservation and refresh problem, leading to PReM, a context-compression framework that maintains long context as internal KV memory without an external compressor.
    \item PReM uses a dedicated memory layer and a special memory token \(\langle m\rangle\) to dynamically select, preserve, and refresh context memory, and introduces phase-separated refresh training to align step-specific memory selection with memory-conditioned generation.
    \item Experiments with 32K-token contexts show that PReM outperforms strong KV-cache and context-compression baselines under 16× and 32× compression while maintaining favorable inference efficiency.
\end{itemize}

\section{Related Work}
\paragraph{KV-Cache Compression}
KV-cache compression reduces the memory and attention cost of autoregressive inference by retaining only part of the cached KV states.
Existing methods exploit different properties of the cache, such as heavy-hitter tokens~\cite{zhang2023h2oheavyhitteroracleefficient}, attention sinks~\cite{ICLR2024_streamingllm}, layer-wise cache allocation~\cite{yang2024pyramidinferpyramidkvcache,cai2025pyramidkvdynamickvcache}, recent attention patterns~\cite{NEURIPS2024_snapkv}, or adaptive cache eviction~\cite{qin2025cake,yu-chai-2025-evolkv,li2026intentkvcrossturnintentawarekv}.
These methods mainly operate on the cache produced during inference and aim to preserve useful past states under a fixed memory budget.
PReM shares the goal of reducing KV memory cost, but differs in two aspects: it compresses an external long-context memory represented as layer-wise KV states, and it learns step-specific memory selection from the model's internal hidden states rather than relying only on generic cache heuristics.

\paragraph{Context Compression}
Context-compression methods reduce long-context cost by shortening the input before generation.
Text-space compressors select or rewrite informative tokens and passages before feeding them to the downstream model~\cite{jiang-etal-2024-longllmlingua,pan-etal-2024-llmlingua,hwang2025exitcontextawareextractivecompression,zhao-etal-2025-dac,chirkova2025provence}, while soft-memory methods encode the context into compact learned tokens or activations~\cite{mu2023gisttokens,ICLR2024_ICAE,ICLR2025_Activation_Beacon,li-etal-2025-500xcompressor,yu2026srki,tang2026comicoarsetofinecontextcompression}.
These approaches can substantially reduce the effective input length, but many of them perform compression once before generation or depend on a separate compressor, which can separate the compressed representation from the model's evolving generation state.
PReM instead keeps context memory inside the LLM as KV states and refreshes it through the generated memory token \(\langle m\rangle\).
This allows compression to be conditioned on the current reasoning step and trained jointly with memory-conditioned generation.

\begin{figure*}[htbp]
    \centering

    \begin{minipage}[t]{0.9\textwidth}
        \centering
        \vspace{0pt}
        \includegraphics[width=\linewidth]{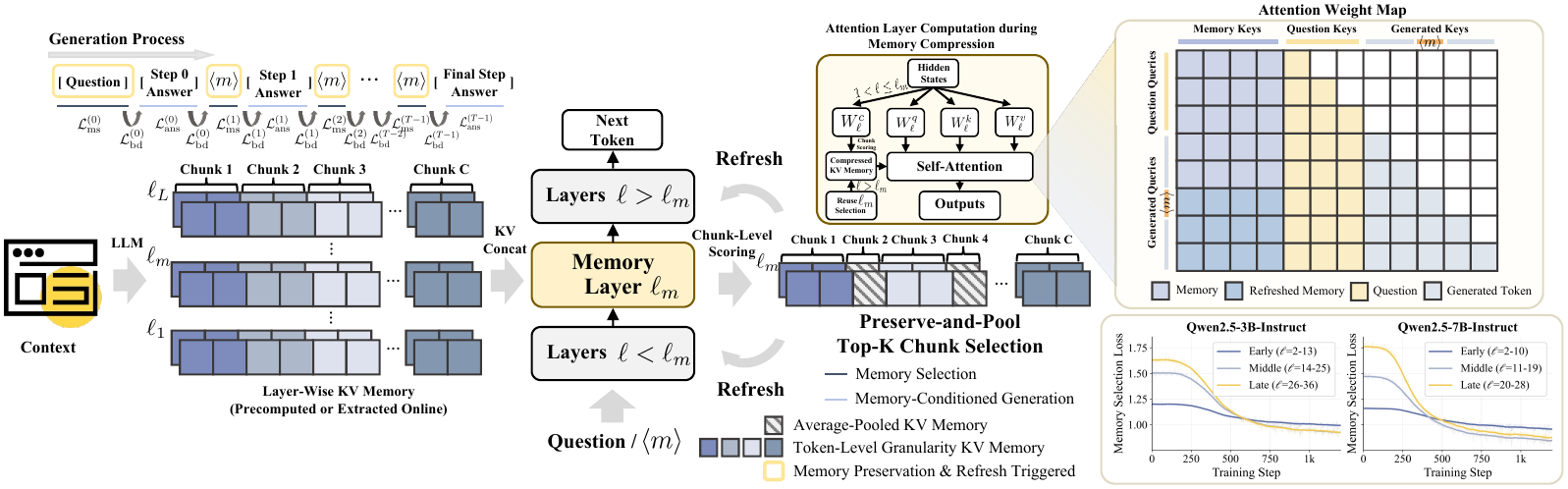}
    \end{minipage}

    \caption{\textbf{Left:} memory preservation and refresh in PReM. A dedicated memory layer selects evidence chunks, and PReM reuses their indices to refresh memory across other layers. When the model emits the \(\langle m\rangle\), the memory is reselected and refreshed. Here, \(W_{\ell}^{c}\) denotes the selection query projection matrix, while \(W_{\ell}^{q}\), \(W_{\ell}^{k}\), and \(W_{\ell}^{v}\) denote the standard query, key, and value projection matrices. \textbf{Top right:} compressed KV memory is incorporated into standard attention, while \(\langle m\rangle\) triggers memory refresh.
\textbf{Bottom right:} average memory selection loss over each layer range during 4K-context training in Qwen models.}
    \label{fig:rrem_overview_and_loss}
\end{figure*}

\section{Preserve-and-Refresh Context Compression}

\subsection{Motivation}
\label{sec:motivation}

Different reasoning steps may depend on different contextual evidence, so an initially selected memory can become stale as generation proceeds.
PReM therefore uses the hidden state of \(\langle m\rangle\) to refresh memory during generation.
Because this refresh operates over layer-wise KV memory whose internal representations vary across depths, this raises a second design question: \textit{which layer provides a more suitable representation for memory selection?}
We answer this with a layer-wise diagnostic that applies the memory selection loss defined in Section~\ref{sec:stepwise-training} to each candidate layer; this loss trains the model to identify evidence chunks needed at each reasoning step.
Across Qwen~\cite{qwen2025qwen25technicalreport} backbones of different scales, as shown in Figure~\ref{fig:rrem_overview_and_loss} (bottom right), this loss decreases unevenly: middle and late layers often learn memory selection faster, while early layers exhibit slower loss reduction.
This suggests that layer-wise KV memory is not equally suitable for evidence selection at all depths, motivating PReM to designate a middle-to-late layer as the \emph{memory layer} to control memory selection.

\subsection{PReM}
As shown in Figure~\ref{fig:rrem_overview_and_loss} (left), PReM formulates context compression as learned updates over layer-wise context memory stored as cached KV states in the LLM: the model decides what to preserve and when to refresh it.
PReM realizes this with two internal mechanisms: a dedicated memory layer that controls memory selection, and a special memory token \(\langle m\rangle\) that serves as the refresh signal.

\paragraph{Context Memory}

Before generation, PReM encodes the long context into a layer-wise KV memory.
For an LLM with \(L\) layers and \(H\) attention heads, the KV memory at layer \(\ell\) is represented as \(M_{\ell}=(K_{\ell}^{M},V_{\ell}^{M})\), where \(\ell \in \{1,\ldots,L\}\), \(K_{\ell}^{M},V_{\ell}^{M}\in\mathbb{R}^{H\times N\times d}\), \(N\) is the number of context tokens, and \(d\) is the head dimension.
In implementation, these memory states can be loaded from precomputed context KV states or extracted online from the current model.
The KV memory will later be compressed and prepended to the KV states of the current sequence, participating in standard attention as additional key-value states, as shown in Figure~\ref{fig:rrem_overview_and_loss} (top right).

\paragraph{Memory Selection and Refresh}

Memory selection is performed at chunk granularity.
Specifically, given a chunk size \(c\), PReM partitions the \(N\) context memory positions into \(C=\lceil N/c\rceil\) contiguous chunks, where the \(j\)-th chunk is \(\mathcal{C}_{j}=\{(j-1)c+1,\ldots,\min(jc,N)\}\), for \(j\in\{1,\ldots,C\}\).
To learn this selection from internal model states, PReM uses a dedicated memory layer \(\ell_m\) and adds a selection query projection matrix \(W_{\ell}^{c}\) to layers \(1<\ell\le \ell_m\).
This projection maps the attention-layer input hidden state to a selection query with the same shape as the standard attention query, which is later used to score chunks for memory selection.
Layers above \(\ell_m\) do not use \(W_{\ell}^{c}\); they reuse the memory selection produced by the memory layer.
The embedding-adjacent layer \(\ell=1\) is excluded because the \(\langle m\rangle\) representation is still an embedding-level signal there; using it for memory selection would make the selected chunks nearly fixed rather than refresh-dependent.
This layer instead provides an uncompressed full-context access path.

Given the question \(x\), PReM organizes generation into \(T\) reasoning steps:
\([x,y_0,\langle m\rangle,y_1,\langle m\rangle,\ldots,y_{T-1}]\), where \(y_s\) is the answer segment generated at step \(s\), for \(s \in \{0,\ldots,T-1\}\).
Step \(0\) selects memory using selection queries derived from \(x\); for \(s>0\), the \(\langle m\rangle\) token before \(y_s\) triggers a refresh and provides the selection query for the current step.
Let \(\mathcal{P}_{s}\) denote the corresponding query positions.
For chunk-level scoring at layer \(\ell\), PReM first averages key states over the token positions within each chunk:
\(\bar{K}_{\ell,j}^{M}=|\mathcal{C}_{j}|^{-1}\sum_{t\in\mathcal{C}_{j}}K_{\ell,t}^{M}\) and \(\bar{k}_{\ell,h,j}^{M}\) denotes its head-\(h\) slice for \(h\in\{1,\ldots,H\}\).
Given the step-\(s\) selection queries \(q_{\ell,h,p}^{(s)}\) at positions \(p\in\mathcal{P}_{s}\), PReM then computes an attention-like score for each chunk:
\begin{equation}
a_{\ell,s,j}
= \frac{1}{H|\mathcal{P}_{s}|}
\sum_{h=1}^{H}
\sum_{p \in \mathcal{P}_{s}}
\frac{
\left(q_{\ell,h,p}^{(s)}\right)^{\top}
\bar{k}_{\ell,h,j}^{M}
}{\sqrt{d}} .
\end{equation}
At layer \(\ell_m\), these scores produce the authoritative selection
\(\mathcal{S}_{s}=\operatorname{TopK}(\{a_{\ell_m,s,j}\}_{j=1}^{C}, k)\), where \(k\) is the chunk budget.

Given \(\mathcal{S}_{s}\), PReM constructs the compressed memory with a preserve-and-pool rule.
For \(j\in\mathcal{S}_{s}\), it preserves all token-level KV states in the chunk, \(\widetilde{M}_{\ell,j}^{(s)}=\{(K_{\ell,t}^{M},V_{\ell,t}^{M}):t\in\mathcal{C}_{j}\}\).
For \(j\notin\mathcal{S}_{s}\), it replaces the chunk with one representative pair, \(\widetilde{M}_{\ell,j}^{(s)}=\{(\bar{K}_{\ell,j}^{M},\bar{V}_{\ell,j}^{M})\}\), where \(\bar{V}_{\ell,j}^{M}=|\mathcal{C}_{j}|^{-1}\sum_{t\in\mathcal{C}_{j}}V_{\ell,t}^{M}\).
At each step, the original token-level memory remains available as the source for future refreshes, while the compressed memory used for subsequent generation is replaced with a newly constructed one.
Because selection is performed at chunk granularity and each refresh is triggered by a single \(\langle m\rangle\) token, this refresh remains lightweight.
For position encoding, selected chunks keep their original context positions, while an unselected chunk uses its center position \(p_j=(j-1)c+\lfloor(|\mathcal{C}_{j}|-1)/2\rfloor\); rotary position encoding is then applied with these positions.

To propagate the memory-layer decision across depths, PReM reuses the selection \(\mathcal{S}_{s}\) across layers.
Before the computation reaches \(\ell_m\), the lower layers with selection query projections temporarily apply the compression process described above, reducing the cost of the refresh forward pass.
Once \(\mathcal{S}_{s}\) is produced, the memory in the lower layers is refreshed with the memory-layer selection.
Layers above \(\ell_m\) skip chunk scoring and directly compress their own KV memory using \(\mathcal{S}_{s}\).
Through this design, these layers share the same memory selection at each step when generating the corresponding answer segment.
We denote the resulting layer-wise compressed memory used during step \(s\) by \(\widetilde{M}^{(s)}\).

\paragraph{Phase-Separated Refresh Training}
\label{sec:stepwise-training}

Refresh training follows the step structure \([x,y_0,\langle m\rangle,y_1,\langle m\rangle,\ldots,y_{T-1}]\) defined above, using teacher forcing over this target sequence.
PReM then decomposes each step into two phases with separate forward passes: a \emph{memory-selection phase} and a \emph{memory-conditioned generation phase}, as shown in Figure~\ref{fig:rrem_overview_and_loss} (top left).
Let \(z_0=x\) and \(z_s=\langle m\rangle\) for \(s>0\).
The selection phase processes only \(z_s\), computes \(\mathcal{S}_{s}\), optimizes the memory selection loss, and refreshes all layers that maintain context memory using the reuse rule above. The generation phase then processes the teacher-forced segment \(y_s\) after this refresh.
This separation is necessary for training-inference consistency.
At inference time, each refresh changes the KV memory used by subsequent tokens, so the answer tokens in each step should be generated under the newly refreshed memory.
Accordingly, the training forward pass separates these roles: \(z_s\) performs selection and refresh, while \(y_s\) is generated under the resulting memory state.

We next specify the memory selection loss used in the selection phase.
For each step \(s\), let \(\mathcal{G}_{s}\subseteq\{1,\ldots,C\}\) denote the evidence chunk indices needed to generate \(y_s\).
The loss trains the memory layer \(\ell_m\) to rank evidence chunks above non-evidence chunks using the chunk scores \(\{a_{\ell_m,s,j}\}_{j=1}^{C}\).
Let \(\tau\) be the softmax temperature.
For multiple evidence chunks, we average the one-positive contrastive losses while excluding the other positives from the denominator:
\begin{equation}
\mathcal{L}_{\mathrm{ms}}^{(s)}
= -\frac{1}{|\mathcal{G}_{s}|}
\sum_{g\in\mathcal{G}_{s}}
\log
\frac{e^{a_{\ell_m,s,g}/\tau}}
{e^{a_{\ell_m,s,g}/\tau}+\sum_{j\notin\mathcal{G}_{s}}e^{a_{\ell_m,s,j}/\tau}}.
\end{equation}
This objective is computed using the same query positions as memory selection: all question positions at \(s=0\) and the \(\langle m\rangle\) position for \(s>0\).

The language modeling objective must account for the boundaries introduced by the two-phase forward pass.
Write the answer segment as \(y_s=(y_{s,1},\ldots,y_{s,n_s})\), where \(n_s\) is the number of tokens in segment \(y_s\).
The generation phase gives the ordinary within-segment cross entropy:
\begin{equation}
\mathcal{L}_{\mathrm{ans}}^{(s)}
= -\sum_{r=1}^{n_s-1}
\log p(y_{s,r+1}\mid z_{\le s}, y_{<s}, y_{s,\le r}, \widetilde{M}^{(s)}).
\end{equation}
However, because the selection phase and generation phase are forwarded separately, two autoregressive links would otherwise be missing: the link from \(z_s\) to the first answer token, and the link from the end of \(y_s\) to the next memory token.
PReM therefore adds boundary cross entropy terms:
\begin{equation}
\begin{aligned}
\mathcal{L}_{\mathrm{bd}}^{(s)}
=& -\log p(y_{s,1}\mid z_{\le s}, y_{<s}, \widetilde{M}^{(s)}) \\
& - \mathbf{1}_{s<T-1}
\log p(z_{s+1}\mid z_{\le s}, y_{\le s}, \widetilde{M}^{(s)}).
\end{aligned}
\end{equation}
These boundary terms are not assigned a separate loss weight; they are merged into the normal language-modeling cross entropy so that the segmented training has the same token-level coverage as an unsegmented autoregressive pass.
The second boundary term also trains the model to emit \(\langle m\rangle\) when the next step should refresh memory.
The full training objective is
\begin{equation}
\mathcal{L}
= \frac{1}{N_{\mathrm{LM}}}
\sum_{s=0}^{T-1}
\left(\mathcal{L}_{\mathrm{ans}}^{(s)}+\mathcal{L}_{\mathrm{bd}}^{(s)}\right)
+ \lambda_{\mathrm{ms}}
\frac{1}{T}
\sum_{s=0}^{T-1}
\mathcal{L}_{\mathrm{ms}}^{(s)},
\end{equation}
where \(N_{\mathrm{LM}}\) is the number of supervised language-modeling tokens and \(\lambda_{\mathrm{ms}}\) controls the weight of memory selection supervision.

\paragraph{Inference}

Before answer generation, PReM constructs the initial compressed memory using selection queries from the input question tokens.
The resulting compressed KV memory is used in the standard attention computation together with the current sequence states.
During decoding, ordinary tokens keep using the current compressed memory cached inside each layer.
When the model generates \(\langle m\rangle\), PReM performs a refresh step: the memory layer uses the selection query of \(\langle m\rangle\) to score all chunks and select a new top-\(k\) set.
Selected chunks are preserved, unselected chunks are average-pooled, and the resulting selection refreshes the compressed memory across layers.
This procedure gives the model two degrees of freedom under a fixed memory budget: it decides \emph{what} to preserve by selecting chunks at each refresh step, and \emph{when} to refresh by generating \(\langle m\rangle\) whenever later reasoning requires different contextual evidence.
PReM therefore implements context compression as a sequence of learned memory updates rather than a single pre-generation shortening operation.

\begin{table*}[t]
\centering
\scriptsize
\setlength{\tabcolsep}{5.8pt}
\begin{tabular}{clrrrrrrrrrrrrrr}
\toprule
\multirow{2}{*}{\textbf{Model}} & \multirow{2}{*}{\textbf{Method}} & \multicolumn{2}{c}{\textbf{TriviaQA}} & \multicolumn{2}{c}{\textbf{SQuAD}} & \multicolumn{2}{c}{\textbf{NaturalQuestions}} & \multicolumn{2}{c}{\textbf{2WikiMQA}} & \multicolumn{2}{c}{\textbf{HotpotQA}} & \multicolumn{2}{c}{\textbf{MuSiQue}} & \multicolumn{2}{c}{\textbf{Avg.}} \\
\cmidrule(lr){3-4}\cmidrule(lr){5-6}\cmidrule(lr){7-8}\cmidrule(lr){9-10}\cmidrule(lr){11-12}\cmidrule(lr){13-14}\cmidrule(lr){15-16}
 & & \textbf{EM} & \textbf{F1} & \textbf{EM} & \textbf{F1} & \textbf{EM} & \textbf{F1} & \textbf{EM} & \textbf{F1} & \textbf{EM} & \textbf{F1} & \textbf{EM} & \textbf{F1} & \textbf{EM} & \textbf{F1} \\
\hline
\multicolumn{16}{l}{\textbf{Qwen2.5-3B-Instruct}} \\ 
\multicolumn{2}{l}{\hspace{1.8em}Prompting (no-context)} & 0.00 & 0.29 & 0.00 & 0.26 & 3.60 & 5.70 & 5.60 & 6.93 & 3.80 & 5.74 & 0.80 & 3.04 & 2.30 & 3.66 \\
\multicolumn{2}{l}{\hspace{1.8em}Prompting} & 10.80 & 15.86 & 10.60 & 15.82 & 9.60 & 14.53 & 17.40 & 22.90 & 25.00 & 34.07 & 8.40 & 12.15 & 13.63 & 19.22 \\ \hline
\multicolumn{16}{l}{\textbf{Qwen2.5-7B-Instruct}} \\ 
\multicolumn{2}{l}{\hspace{1.8em}Prompting (no-context)} & 0.20 & 0.43 & 0.00 & 0.09 & 0.40 & 0.49 & 0.00 & 0.00 & 0.00 & 0.00 & 0.00 & 0.14 & 0.10 & 0.19 \\
\multicolumn{2}{l}{\hspace{1.8em}Prompting} & 21.40 & 24.29 & 18.60 & 21.91 & 20.60 & 25.81 & 23.00 & 28.66 & 31.00 & 39.96 & 16.80 & 21.86 & 21.90 & 27.08 \\
\hline
\multicolumn{16}{c}{\textbf{16× compression constraint}} \\
\hline
\multirow{7}{*}{\rotatebox[origin=c]{90}{\textbf{Qwen2.5-3B-Instruct}}} & SnapKV & \textbf{30.00} & \textbf{36.58} & \underline{25.00} & \underline{32.41} & 9.20 & 14.35 & 10.60 & 13.04 & 20.20 & 27.55 & 4.40 & 6.75 & 16.57 & 21.78 \\
& StreamingLLM & 9.40 & 12.47 & 4.20 & 6.35 & 5.80 & 7.00 & 7.20 & 7.83 & 9.80 & 14.12 & 4.00 & 6.48 & 6.73 & 9.04 \\
& CAKE & \underline{29.40} & \underline{36.03} & 23.60 & 31.42 & \underline{12.20} & \underline{16.81} & 10.40 & 14.78 & 20.40 & 28.56 & 6.40 & 10.17 & 17.07 & 22.96 \\
& LongLLMLingua & 25.40 & 28.77 & 17.20 & 21.95 & 11.20 & 15.42 & \underline{12.20} & 14.47 & 21.00 & 28.31 & 7.00 & 9.45 & 15.67 & 19.73 \\
& LLMLingua-2-large & 10.80 & 14.37 & 4.00 & 6.35 & 3.40 & 4.80 & 3.40 & 4.51 & 6.20 & 9.07 & 3.40 & 5.32 & 5.20 & 7.40 \\
& EXIT & 25.80 & 31.83 & \textbf{28.20} & \textbf{32.48} & 11.20 & 15.01 & 11.20 & \underline{14.94} & \textbf{35.20} & \textbf{45.87} & \underline{15.20} & \underline{20.04} & \underline{21.13} & \underline{26.70} \\
& \premcell{PReM} & \premcell{20.40} & \premcell{24.89} & \premcell{17.40} & \premcell{21.21} & \premcell{\textbf{32.80}} & \premcell{\textbf{37.69}} & \premcell{\textbf{37.40}} & \premcell{\textbf{42.66}} & \premcell{\underline{30.40}} & \premcell{\underline{39.61}} & \premcell{\textbf{20.20}} & \premcell{\textbf{25.05}} & \premcell{\textbf{26.43}} & \premcell{\textbf{31.85}} \\ \hline
\multirow{7}{*}{\rotatebox[origin=c]{90}{\textbf{Qwen2.5-7B-Instruct}}} & SnapKV & 27.20 & 31.47 & 21.00 & 25.98 & \underline{21.80} & \underline{27.47} & 14.60 & \underline{20.55} & 27.00 & 37.91 & 10.00 & 15.58 & 20.27 & 26.49 \\
& StreamingLLM & 15.80 & 19.15 & 8.20 & 11.77 & 14.80 & 17.78 & 11.00 & 12.23 & 13.00 & 17.63 & 4.40 & 7.67 & 11.20 & 14.37 \\
& CAKE & 28.00 & 32.24 & 19.20 & 22.94 & 20.80 & 26.86 & 14.80 & 20.29 & 28.40 & 38.36 & 9.80 & 15.76 & 20.17 & 26.08 \\
& LongLLMLingua & \textbf{33.60} & \textbf{38.58} & 25.60 & \underline{30.54} & 14.00 & 17.23 & \underline{15.20} & 18.07 & 26.80 & 34.51 & 11.00 & 16.43 & 21.03 & 25.89 \\
& LLMLingua-2-large & 14.60 & 17.23 & 4.80 & 6.99 & 6.20 & 7.60 & 1.80 & 2.67 & 4.80 & 6.71 & 5.20 & 6.98 & 6.23 & 8.03 \\
& EXIT & \underline{31.40} & 35.89 & \textbf{27.00} & 30.01 & 13.20 & 17.61 & 12.20 & 15.12 & \underline{38.80} & \underline{48.94} & \underline{24.20} & \underline{29.43} & \underline{24.47} & \underline{29.50} \\
& \premcell{PReM} & \premcell{31.00} & \premcell{\underline{36.30}} & \premcell{\underline{26.40}} & \premcell{\textbf{31.35}} & \premcell{\textbf{38.80}} & \premcell{\textbf{47.04}} & \premcell{\textbf{45.80}} & \premcell{\textbf{51.96}} & \premcell{\textbf{39.40}} & \premcell{\textbf{52.37}} & \premcell{\textbf{26.80}} & \premcell{\textbf{33.26}} & \premcell{\textbf{34.70}} & \premcell{\textbf{42.05}} \\
\hline
\multicolumn{16}{c}{\textbf{32× compression constraint}} \\
\hline
\multirow{7}{*}{\rotatebox[origin=c]{90}{\textbf{Qwen2.5-3B-Instruct}}} & SnapKV & \textbf{33.00} & \textbf{38.86} & 22.40 & 29.97 & 7.60 & 11.77 & 7.80 & 9.21 & 17.20 & 24.55 & 5.00 & 8.00 & 15.50 & 20.39 \\
& StreamingLLM & 9.00 & 12.35 & 2.60 & 4.86 & 5.20 & 6.90 & 8.00 & 8.30 & 7.20 & 10.31 & 2.40 & 4.46 & 5.73 & 7.86 \\
& CAKE & 31.00 & \underline{38.14} & \underline{24.20} & \underline{31.43} & 11.20 & 16.19 & 9.20 & \underline{12.51} & 19.40 & 27.61 & 5.20 & 8.09 & 16.70 & 22.33 \\
& LongLLMLingua & 24.40 & 29.13 & 12.20 & 16.41 & 8.40 & 12.75 & \underline{9.60} & 11.46 & 14.80 & 22.25 & 5.20 & 7.35 & 12.43 & 16.56 \\
& LLMLingua-2-large & 9.20 & 11.88 & 3.20 & 4.83 & 4.80 & 5.84 & 2.60 & 3.47 & 5.60 & 7.77 & 2.60 & 3.87 & 4.67 & 6.28 \\
& EXIT & \underline{31.80} & 37.56 & \textbf{30.00} & \textbf{35.08} & \underline{14.40} & \underline{18.90} & 8.80 & 11.37 & \textbf{31.00} & \textbf{39.25} & \underline{14.40} & \underline{19.21} & \underline{21.73} & \underline{26.90} \\
& \premcell{PReM} & \premcell{16.60} & \premcell{22.25} & \premcell{19.00} & \premcell{22.48} & \premcell{\textbf{33.20}} & \premcell{\textbf{39.51}} & \premcell{\textbf{35.00}} & \premcell{\textbf{39.62}} & \premcell{\underline{26.80}} & \premcell{\underline{36.18}} & \premcell{\textbf{18.20}} & \premcell{\textbf{24.15}} & \premcell{\textbf{24.80}} & \premcell{\textbf{30.70}} \\ \hline
\multirow{7}{*}{\rotatebox[origin=c]{90}{\textbf{Qwen2.5-7B-Instruct}}} & SnapKV & 30.80 & 35.76 & 23.40 & 27.64 & \underline{21.40} & \underline{27.87} & 13.40 & 17.52 & 25.20 & 34.01 & 9.60 & 15.16 & 20.63 & 26.33 \\
& StreamingLLM & 17.40 & 20.39 & 7.80 & 11.51 & 12.80 & 15.97 & 8.60 & 10.53 & 11.80 & 16.58 & 4.40 & 8.49 & 10.47 & 13.91 \\
& CAKE & 31.80 & 36.53 & 22.40 & 26.77 & 20.20 & 27.33 & \underline{14.20} & \underline{18.09} & 25.00 & 33.75 & 8.00 & 14.57 & 20.27 & 26.17 \\
& LongLLMLingua & \underline{32.20} & \underline{36.71} & 23.40 & 28.07 & 14.60 & 19.51 & 12.40 & 15.66 & 25.00 & 31.26 & 9.60 & 13.54 & 19.53 & 24.13 \\
& LLMLingua-2-large & 12.00 & 14.28 & 2.40 & 3.72 & 4.00 & 5.31 & 1.20 & 1.25 & 4.20 & 4.77 & 4.20 & 5.25 & 4.67 & 5.76 \\
& EXIT & \textbf{38.80} & \textbf{43.77} & \textbf{33.60} & \textbf{37.86} & 13.80 & 17.72 & 9.80 & 12.30 & \underline{35.80} & \underline{45.49} & \underline{17.40} & \underline{23.93} & \underline{24.87} & \underline{30.18} \\
& \premcell{PReM} & \premcell{25.60} & \premcell{32.04} & \premcell{\underline{25.20}} & \premcell{\underline{29.55}} & \premcell{\textbf{39.80}} & \premcell{\textbf{46.68}} & \premcell{\textbf{42.20}} & \premcell{\textbf{48.59}} & \premcell{\textbf{37.60}} & \premcell{\textbf{48.39}} & \premcell{\textbf{28.00}} & \premcell{\textbf{35.19}} & \premcell{\textbf{33.07}} & \premcell{\textbf{40.07}} \\
\bottomrule
\end{tabular}
\caption{Main results using the 3B and 7B variants of Qwen2.5-Instruct as backbones. ``Prompting'' denotes feeding the full 32K-token context to the model without compression, while ``no-context'' denotes direct answering without any context. Best and second-best results per model/compression block are bold and underlined. ``Avg.'' denotes average scores.}
\label{tab:main-results}
\end{table*}

\begin{table*}[t]
\centering
\scriptsize
\setlength{\tabcolsep}{6.5pt}
\begin{tabular}{lrrrrrrrrrrrrrr}
\toprule
\multirow{2}{*}{\textbf{Method}} & \multicolumn{2}{c}{\textbf{TriviaQA}} & \multicolumn{2}{c}{\textbf{SQuAD}} & \multicolumn{2}{c}{\textbf{NaturalQuestions}} & \multicolumn{2}{c}{\textbf{2WikiMQA}} & \multicolumn{2}{c}{\textbf{HotpotQA}} & \multicolumn{2}{c}{\textbf{MuSiQue}} & \multicolumn{2}{c}{\textbf{Avg.}} \\
\cmidrule(lr){2-3}\cmidrule(lr){4-5}\cmidrule(lr){6-7}\cmidrule(lr){8-9}\cmidrule(lr){10-11}\cmidrule(lr){12-13}\cmidrule(lr){14-15}
 & \textbf{EM} & \textbf{F1} & \textbf{EM} & \textbf{F1} & \textbf{EM} & \textbf{F1} & \textbf{EM} & \textbf{F1} & \textbf{EM} & \textbf{F1} & \textbf{EM} & \textbf{F1} & \textbf{EM} & \textbf{F1} \\
\hline
\multicolumn{15}{c}{\textbf{16× compression constraint}} \\
\hline
Activation Beacon & \textbf{37.00} & \textbf{43.75} & \underline{20.40} & \underline{27.37} & 14.40 & 20.80 & 12.20 & 16.06 & 19.60 & 26.96 & 7.40 & 11.20 & 18.50 & 24.36 \\
ICAE & 12.80 & 17.40 & 1.80 & 4.76 & 1.60 & 3.93 & 3.40 & 3.87 & 2.80 & 5.09 & 0.40 & 1.94 & 3.80 & 6.17 \\
\rowcolor{premrow} PReM-3B & 20.40 & 24.89 & 17.40 & 21.21 & \underline{32.80} & \underline{37.69} & \underline{37.40} & \underline{42.66} & \underline{30.40} & \underline{39.61} & \underline{20.20} & \underline{25.05} & \underline{26.43} & \underline{31.85} \\
\rowcolor{premrow} PReM-7B & \underline{31.00} & \underline{36.30} & \textbf{26.40} & \textbf{31.35} & \textbf{38.80} & \textbf{47.04} & \textbf{45.80} & \textbf{51.96} & \textbf{39.40} & \textbf{52.37} & \textbf{26.80} & \textbf{33.26} & \textbf{34.70} & \textbf{42.05} \\
\hline
\multicolumn{15}{c}{\textbf{32× compression constraint}} \\
\hline
Activation Beacon & \textbf{34.60} & \textbf{40.42} & 14.40 & 19.20 & 13.40 & 18.65 & 10.80 & 13.63 & 16.60 & 23.61 & 7.00 & 12.04 & 16.13 & 21.26 \\
ICAE & 12.00 & 17.94 & 1.20 & 4.48 & 1.40 & 3.63 & 3.00 & 3.65 & 2.40 & 4.95 & 0.80 & 2.23 & 3.47 & 6.15 \\
\rowcolor{premrow} PReM-3B & 16.60 & 22.25 & \underline{19.00} & \underline{22.48} & \underline{33.20} & \underline{39.51} & \underline{35.00} & \underline{39.62} & \underline{26.80} & \underline{36.18} & \underline{18.20} & \underline{24.15} & \underline{24.80} & \underline{30.70} \\
\rowcolor{premrow} PReM-7B & \underline{25.60} & \underline{32.04} & \textbf{25.20} & \textbf{29.55} & \textbf{39.80} & \textbf{46.68} & \textbf{42.20} & \textbf{48.59} & \textbf{37.60} & \textbf{48.39} & \textbf{28.00} & \textbf{35.19} & \textbf{33.07} & \textbf{40.07} \\
\bottomrule
\end{tabular}
\caption{Results for released-backbone soft context-compression baselines and PReM. PReM-3B and PReM-7B denote the 3B and 7B backbones of Qwen2.5-Instruct.}
\label{tab:other-backbone-baselines}
\end{table*}

\section{Experiments}

\subsection{Experimental Setup}

\paragraph{Models and Datasets}
We train PReM on the 3B and 7B variants of Qwen2.5-Instruct~\cite{qwen2025qwen25technicalreport}. The main evaluation covers six benchmarks spanning single-hop and multi-hop settings, which require different numbers of reasoning steps and supporting evidence pieces: TriviaQA~\cite{joshi-etal-2017-triviaqa}, SQuAD~\cite{rajpurkar-etal-2016-squad}, NaturalQuestions~\cite{liu-etal-2024-lost-in-middle}, 2WikiMQA~\cite{ho-etal-2020-constructing-2wikimqa}, HotpotQA~\cite{yang-etal-2018-hotpotqa}, and MuSiQue~\cite{trivedi-etal-2022-musique}.
Each dataset is evaluated on 500 examples sampled from its test or development set, and we report exact match (EM) and token-level F1.

\paragraph{Training}
We describe the training-data construction process in Appendix~\ref{app:training-data-construction}. To build this dataset, we annotate the training splits of NaturalQuestions, 2WikiMQA, HotpotQA, and MuSiQue, yielding 32,319 examples in total. Detailed dataset statistics are provided in Appendix~\ref{app:training_data_statistics}. We add \(W_{\ell}^{c}\) to layers \(1<\ell\le \ell_m\) and perform full-parameter training on eight NVIDIA H20 (96GB) GPUs with a 32K-token context, top-\(k=10\), a chunk size of 100, $\tau=0.05$, and $\lambda_{\mathrm{ms}}=0.3$. For memory layer selection, we use layer 32 for Qwen2.5-3B and layer 25 for Qwen2.5-7B. We use DeepSpeed ZeRO Stage 2~\cite{rajbhandari2020zeromemoryoptimizationstraining} with bf16 precision. The learning rate is set to \(1\times10^{-6}\), with a per-device batch size of 1, gradient accumulation steps of 10, a warm-up ratio of \(1\times10^{-2}\), and a weight decay of \(1\times10^{-4}\).

\paragraph{Baselines}
We compare PReM with baselines covering both KV-cache compression and context compression.
\textbf{StreamingLLM}~\cite{ICLR2024_streamingllm} preserves the initial cache and a recent sliding window, forming a fixed-position KV-cache baseline.
\textbf{SnapKV}~\cite{NEURIPS2024_snapkv} selects cached states according to attention from recent query states.
\textbf{CAKE}~\cite{qin2025cake} performs layer-wise KV-cache allocation and importance-based eviction.
\textbf{LongLLMLingua}~\cite{jiang-etal-2024-longllmlingua}, \textbf{LLMLingua-2-large}~\cite{pan-etal-2024-llmlingua}, and \textbf{EXIT}~\cite{hwang-etal-2025-exit} are text-space compression baselines that use separate compressors before generation.
LongLLMLingua and EXIT perform query-aware context compression, while LLMLingua-2-large performs task-agnostic context compression.
We also include two soft context-compression baselines, \textbf{Activation Beacon}~\cite{ICLR2025_Activation_Beacon} with its released Qwen2-7B-Instruct~\cite{yang2024qwen2technicalreport} backbone and \textbf{ICAE}~\cite{ICLR2024_ICAE} with its released Mistral-7B-Instruct-v0.2~\cite{jiang2023mistral7b} backbone, which compress the context into learned memory representations before generation.

\paragraph{Inference Settings}
The main comparison uses a 32K-token context. 
At evaluation time, the long context for each example is constructed from all context documents provided in the benchmark's test or development split.
For PReM, we use a chunk size of 100, with the top-20 chunks for the 16× setting and the top-10 chunks for the 32× setting. For other baselines, we set the target KV cache size or the target number of context tokens to 2K for 16× and 1K for 32×. Detailed settings and prompt templates are provided in Appendix~\ref{app:baseline-inference-settings}.

\subsection{Main Results}

\noindent Table~\ref{tab:main-results} summarizes the Qwen-matched comparison across datasets and compression ratios, while Table~\ref{tab:other-backbone-baselines} reports comparisons with released-backbone baselines.
\noindent\textbf{(1) PReM achieves strong overall performance across compression ratios.}
PReM achieves the best average EM/F1 under both 16× and 32× compression, improving over the strongest Qwen-matched baseline by 5.30/5.15 and 10.23/12.55 points at 16× on 3B/7B, and by 3.07/3.80 and 8.20/9.89 points at 32×.
This shows that learned memory updates remain robust under tighter compression.
\noindent\textbf{(2) PReM is particularly effective on multi-hop QA.}
On multi-hop benchmarks, PReM benefits from step-specific refresh; at 32× on Qwen2.5-7B, it reaches 42.20/48.59 on 2WikiMQA and 28.00/35.19 on MuSiQue, substantially above the strongest Qwen-matched baselines.
These gains suggest that refreshing memory during generation better matches step-wise evidence needs.
We further provide representative case studies in Appendix~\ref{app:qualitative-case-studies}.
\noindent\textbf{(3) Learned compression can outperform direct long-context prompting.}
Even at 32× compression, PReM outperforms direct 32K prompting by 11.17/11.48 points on Qwen2.5-3B and 11.17/12.99 points on Qwen2.5-7B.
Thus, selecting a compact refreshed memory can be more effective than exposing the model to the full context without explicit memory-selection training.
\noindent\textbf{(4) PReM-3B outperforms released 7B soft context-compression baselines.}
As shown in Table~\ref{tab:other-backbone-baselines}, PReM-3B surpasses released 7B soft context-compression baselines, improving over Activation Beacon by 7.93/7.49 points at 16× and 8.67/9.44 points at 32×.
This highlights the strength of internal memory preservation even with a smaller backbone.
Overall, these results show that PReM's gains come not only from reducing context length, but from learning which memory to preserve and when to refresh it.

\subsection{Analyses}

\begin{figure}[t]
    \centering
    \includegraphics[width=0.465\textwidth]{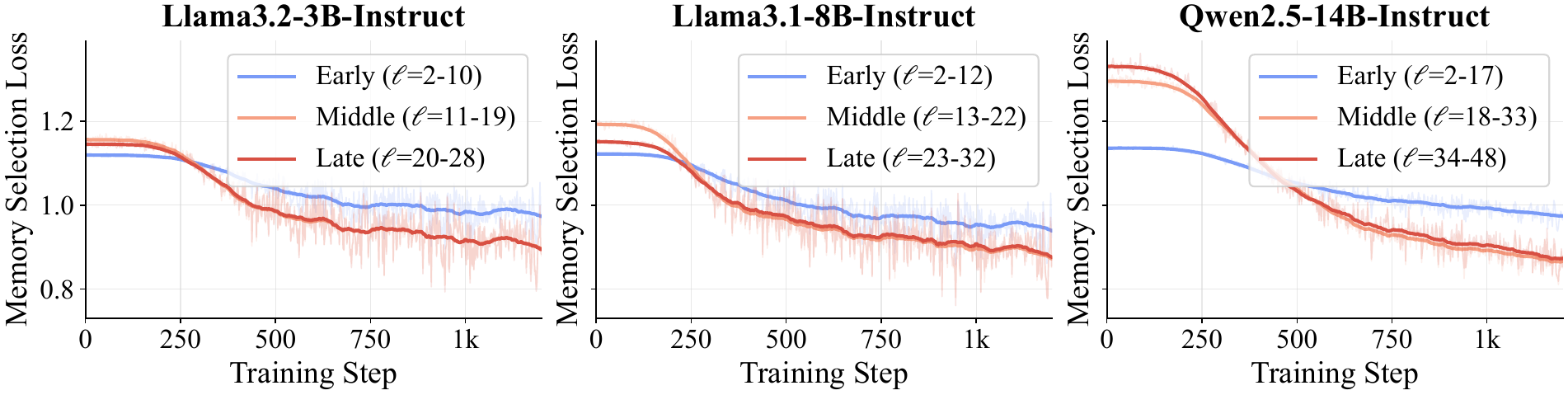}
    \caption{Trends in memory selection loss averaged over the early, middle, and late layer ranges of Llama 3.2-3B-Instruct, Llama 3.1-8B-Instruct, and Qwen2.5-14B-Instruct.
    }
    \label{fig:compression_loss_llama_qwen14}
\end{figure}

\paragraph{Layer-wise memory selection trends generalize beyond Qwen 3B/7B.}
The motivation study in Section~\ref{sec:motivation} shows that memory selection loss decreases unevenly across layers on Qwen backbones.
We further run the same layer-range diagnostic on Llama3.2-3B-Instruct~\cite{meta2024llama32modelcard}, Llama3.1-8B-Instruct~\cite{dubey2024llama}, and Qwen2.5-14B-Instruct~\cite{qwen2025qwen25technicalreport}.
As shown in Figure~\ref{fig:compression_loss_llama_qwen14}, all three models exhibit the same pattern: early layers learn the selection objective more slowly, whereas middle and late layers reduce the loss faster.
This supports our design choice of using a middle-to-late layer as the memory layer for memory selection.

\begin{figure}[tbp]
    \centering
    \includegraphics[width=\linewidth]{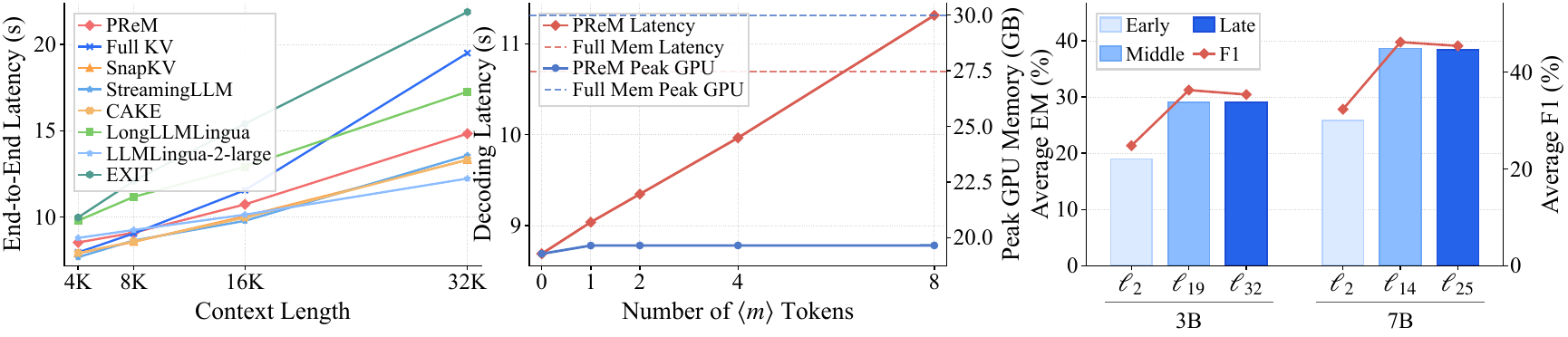}
    \caption{Efficiency and memory-layer analysis of PReM. \textbf{Left}: end-to-end latency vs. context length. \textbf{Middle}: decoding latency and peak GPU memory vs. the number of generated \(\langle m\rangle\) tokens. \textbf{Right}: EM/F1 under early-, middle-, and late-layer choices for the memory layer.
    }
    \label{fig:prem-efficiency}
\end{figure}

\paragraph{PReM scales efficiently with context length and refresh frequency.}
Figure~\ref{fig:prem-efficiency} (left and middle) summarizes the inference cost under a 1K-token context budget and 256-token generation, with chunk size 100 and top-\(k=10\) for PReM.
In these panels, Full KV denotes standard inference with the backbone model, while Full Mem denotes PReM using the full context memory without compression.
All efficiency measurements are conducted on a single NVIDIA H20 GPU using a token-by-token autoregressive decoding loop, and we report the average over 20 runs for each setting.
In the left panel, as context length increases from 4K to 32K, PReM keeps end-to-end latency moderate, reaching 14.83s at 32K compared with 19.50s for Full KV.
External compressors add a separate pre-generation stage, which increases latency for LongLLMLingua and EXIT.
In the middle panel, we evaluate varying numbers of generated \(\langle m\rangle\) tokens under the 32K-token context.
Because Full Mem has no refresh operation, we plot its no-refresh decoding cost as a fixed reference.
With sparse refreshes, PReM decodes in 8.69--9.96s for 0--4 \(\langle m\rangle\) tokens, below the 10.70s Full Mem reference, while more frequent refreshes increase latency.
Meanwhile, decoding peak GPU memory remains nearly constant, averaging 19.57GB, substantially below the 30.0GB Full Mem reference.
Thus \(\langle m\rangle\) exposes a controllable trade-off: the model can spend additional decoding computation to refresh context memory while retaining the memory savings of compression.

\begin{figure}[t]
    \centering
    \includegraphics[width=\linewidth]{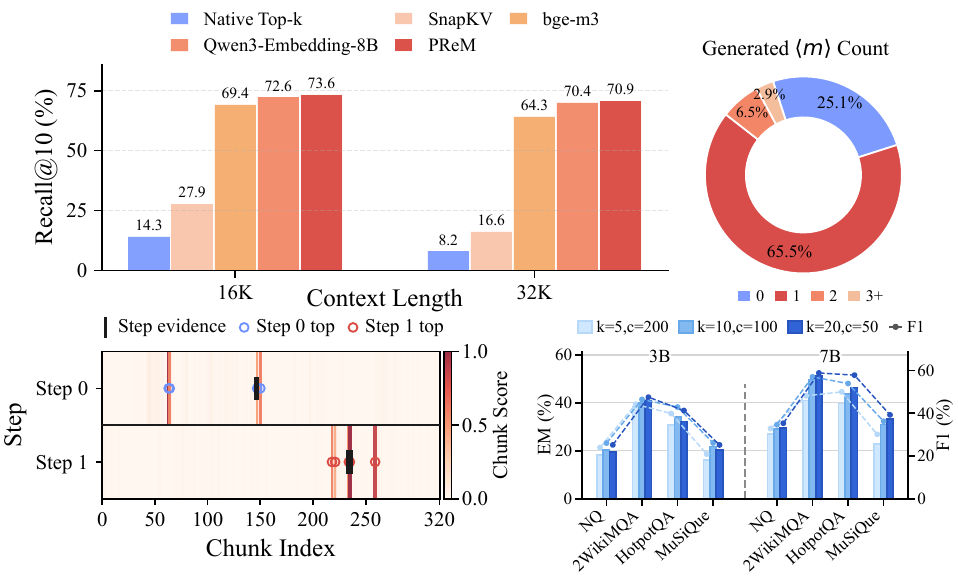}
    \caption{Evidence retrieval, refresh behavior, and chunk-size/top-\(k\) ablations. \textbf{Top left}: micro Recall@10 with 100-token chunks on 128 two-step HotpotQA examples. \textbf{Top right}: generated \(\langle m\rangle\) count distribution. \textbf{Bottom left}: representative 32K case with selected chunks and step evidence marked. \textbf{Bottom right}: chunk-size/top-\(k\) ablation on 3B/7B, reporting EM and F1.}
    \label{fig:recall-attention}
\end{figure}

\paragraph{PReM learns step-specific evidence selection.}
Figure~\ref{fig:recall-attention} first shows, in the top-right donut chart, that PReM uses refresh sparsely in the main-result evaluation: most examples generate one \(\langle m\rangle\) token, while a small fraction generate two or more.
The top-left panel evaluates 128 two-step HotpotQA examples with step-specific required-context annotations, held out from the training set. Using 100-token chunks, we report micro Recall@10. PReM's top-10 selection is scored separately at each step and pooled over all step-level gold chunks; all baselines use a single one-shot selection pass per example and are scored against all gold chunks.
We compare PReM-7B against native top-\(k\), SnapKV, bge-m3~\cite{chen2025m3embeddingmultilingualitymultifunctionalitymultigranularity}, and Qwen3-Embedding-8B~\cite{zhang2025qwen3embeddingadvancingtext}. Native top-\(k\) is training-free and uses the original model attention; for native top-\(k\) and SnapKV, we average layer-wise results. PReM reaches 73.6\%/70.9\% Recall@10 at 16K/32K, substantially outperforming native top-\(k\) and SnapKV and slightly exceeding the strong retriever, Qwen3-Embedding-8B.
The bottom-left qualitative case further shows different answer steps focusing on different evidence chunks.
These results suggest that PReM enables adaptive, precise, step-specific memory access throughout multi-step reasoning.

\subsection{Ablation Studies}

\begin{table}[t]
\centering
\scriptsize
\setlength{\tabcolsep}{2pt}
\begin{tabular}{lrrrrrrrrrr}
\toprule
\multirow{2}{*}{\textbf{Variant}} & \multicolumn{2}{c}{\textbf{NQ}} & \multicolumn{2}{c}{\textbf{2WikiMQA}} & \multicolumn{2}{c}{\textbf{HotpotQA}} & \multicolumn{2}{c}{\textbf{MuSiQue}} & \multicolumn{2}{c}{\textbf{Avg.}} \\
\cmidrule(lr){2-3}\cmidrule(lr){4-5}\cmidrule(lr){6-7}\cmidrule(lr){8-9}\cmidrule(lr){10-11}
 & \textbf{EM} & \textbf{F1} & \textbf{EM} & \textbf{F1} & \textbf{EM} & \textbf{F1} & \textbf{EM} & \textbf{F1} & \textbf{EM} & \textbf{F1} \\
\hline
\multicolumn{11}{c}{\textbf{Qwen2.5-3B-Instruct}} \\ \hline
\rowcolor{premrow} Full & \textbf{20.20} & \textbf{26.02} & \textbf{40.00} & \textbf{46.43} & \textbf{34.20} & \textbf{42.85} & \textbf{21.80} & \textbf{26.08} & \textbf{29.05} & \textbf{35.35} \\
Native top-\(k\) & 7.00 & 10.26 & 4.60 & 6.84 & 7.20 & 9.94 & 2.20 & 4.31 & 5.25 & 7.84 \\
w/o \(\mathcal{L}_{\mathrm{ms}}\) & 7.20 & 10.09 & 13.80 & 16.18 & 9.80 & 15.14 & 7.80 & 11.52 & 9.65 & 13.23 \\
w/o \(\mathcal{L}_{\mathrm{bd}}\) & 18.20 & 23.06 & 18.20 & 21.98 & 18.20 & 23.75 & 5.80 & 8.32 & 15.10 & 19.28 \\
\hline
\multicolumn{11}{c}{\textbf{Qwen2.5-7B-Instruct}} \\ \hline
\rowcolor{premrow} Full & \textbf{29.00} & \textbf{34.59} & \textbf{49.80} & \textbf{56.96} & \textbf{43.60} & \textbf{53.85} & \textbf{31.00} & \textbf{36.20} & \textbf{38.35} & \textbf{45.40} \\
Native top-\(k\) & 13.60 & 19.15 & 7.00 & 9.39 & 18.40 & 23.62 & 4.80 & 8.65 & 10.95 & 15.20 \\
w/o \(\mathcal{L}_{\mathrm{ms}}\) & 18.20 & 23.22 & 22.80 & 25.85 & 20.00 & 27.43 & 13.00 & 19.64 & 18.50 & 24.04 \\
w/o \(\mathcal{L}_{\mathrm{bd}}\) & 25.80 & 30.71 & 27.80 & 32.57 & 30.20 & 39.53 & 14.20 & 17.88 & 24.50 & 30.17 \\
\bottomrule
\end{tabular}
\caption{Ablation results on NQ (NaturalQuestions), 2WikiMQA, HotpotQA, and MuSiQue. ``w/o'' denotes the exclusion of the corresponding component.}
\label{tab:ablation}
\end{table}

\noindent Table~\ref{tab:ablation} summarizes the ablation results. For the trainable variants, we use a 4K-token context, top-\(k=10\), and a chunk size of 100, and train them on the same dataset as in the main results for the same number of training steps.
\noindent\textbf{(1) Explicit memory selection supervision is essential.}
The full PReM model obtains 29.05/35.35 average EM/F1 on Qwen2.5-3B and 38.35/45.40 on Qwen2.5-7B in the ablation setting.
Removing the memory selection loss reduces performance to 9.65/13.23 and 18.50/24.04, respectively, confirming that the memory layer must be directly trained to rank useful context chunks.
\noindent\textbf{(2) Learned chunk selection is not replaced by raw attention.}
Using native attention scores for top-\(k\) selection performs substantially worse, reaching only 5.25/7.84 on 3B and 10.95/15.20 on 7B.
This suggests that ordinary attention over the current sequence is not a reliable proxy for deciding which cached context memory should be preserved.
\noindent\textbf{(3) Boundary modeling matters for refresh training.}
Removing the boundary loss drops performance to 15.10/19.28 on 3B and 24.50/30.17 on 7B.
The degradation shows that the model benefits from explicitly training the autoregressive links between the memory-selection phase, the answer segment, and the next \(\langle m\rangle\) trigger.
\noindent\textbf{(4) Middle and late layers are better suited for memory refresh.}
As shown in Figure~\ref{fig:prem-efficiency} (right), we train PReM with early, middle, and late memory layers under the same settings, using layers 2/19/32 for Qwen2.5-3B and layers 2/14/25 for Qwen2.5-7B. 
We report EM/F1 averaged over the four datasets used in this ablation. Middle and late layers outperform the early setting on both backbones, suggesting that representations from deeper layers provide better control signals for memory refresh. This supports using a middle-to-late layer as the memory layer in our main experiments.
\noindent\textbf{(5) Moderate chunk granularity works best.}
Under the same training setting, we train additional PReM variants with different chunk-size/top-\(k\) choices.
As shown in Figure~\ref{fig:recall-attention} (bottom right), PReM maintains strong performance with 50- and 100-token chunks, while the coarser 200-token setting is consistently weaker.
This suggests that finer chunk granularity helps preserve step-relevant evidence for memory refresh.

\section{Conclusion}

This paper presents PReM, which formulates long-context compression as memory preservation and refresh.
PReM maintains context as layer-wise KV memory, uses a dedicated memory layer and a memory token \(\langle m\rangle\) for step-specific refresh, and trains this behavior with phase-separated refresh training.
Experiments and analyses with 32K-token contexts show that PReM improves answer quality and efficiency under 16× and 32× compression, suggesting that effective context compression should rely on generation-coupled memory updates rather than static context shortening.

\bibliography{aaai2027}
\appendix
\def\mainpaperinput{}
\ifdefined\mainpaperinput
\else
\documentclass[letterpaper]{article}
\usepackage[submission]{aaai2027}
\usepackage[hyphens]{url}
\usepackage{graphicx}
\urlstyle{rm}
\def\UrlFont{\rm}
\usepackage{natbib}
\usepackage{caption}
\frenchspacing

\usepackage{algorithm}
\usepackage{algorithmic}
\usepackage{booktabs}
\usepackage{multirow}
\usepackage{amsmath}
\usepackage{amssymb}
\usepackage[table]{xcolor}
\usepackage{soul}
\usepackage{subcaption}
\sethlcolor{yellow}

\setcounter{secnumdepth}{2}
\pdfinfo{
/TemplateVersion (2027.1)
}

\affiliations{}

\begin{document}

\appendix
\fi

\section{Details of Training Dataset}

\subsection{Training Dataset Construction Details}
\label{app:training-data-construction}

\begin{figure}[h]
\centering
\includegraphics[width=0.44\textwidth]{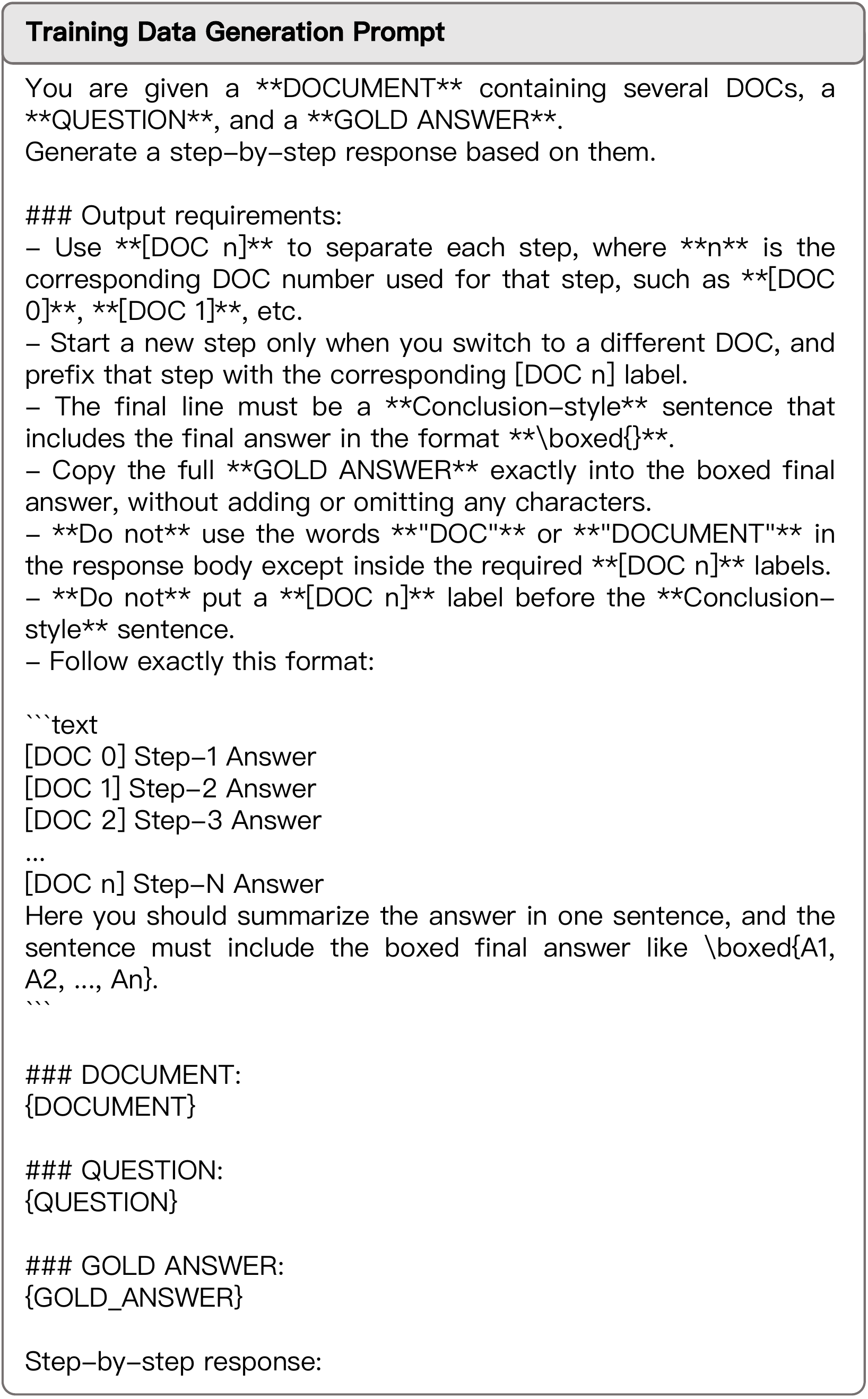}
\caption{Prompt template used to construct document-grounded stepwise training targets.}
\label{fig:training-data-construction-prompt}
\end{figure}

Algorithm~\ref{alg:training-data-construction} summarizes the construction pipeline.
The implementation first normalizes HotpotQA, 2WikiMQA, MuSiQue, and NaturalQuestions into a shared document-indexed schema.
It programmatically filters examples whose answers are missing or not supported by the gold context, then builds global mappings between context text and context ids for later support alignment.
For each instance, the gold context selected by the original support annotations, the question, and the answer are provided to gpt-5.4-2026-03-05
~\cite{openai2026gpt54model} with the construction prompt shown in Figure~\ref{fig:training-data-construction-prompt}.
The prompt asks the annotator model to rewrite the answer into document-grounded segments and to associate each segment with its supporting context.
The parsed segments are aligned back to context ids, yielding step-level support fields stored in the training data.
During training, these stored supports are dynamically mapped to token positions in the context memory and then to the chunk labels \(\mathcal{G}_{s}\) used by the memory selection loss.
Adjacent answer segments are separated by \(\langle m\rangle\), defining the refresh boundary used by stepwise training.
Table~\ref{tab:training-data-examples} provides representative processed examples with different numbers of inserted \(\langle m\rangle\) tokens.
For compactness, the table shows the stepwise target sequence and omits the stored context support fields.

\begin{algorithm}[!h]
\caption{Stepwise Memory-Supervised Data Construction}
\label{alg:training-data-construction}
\scriptsize
\begin{algorithmic}[1]
\REQUIRE
  Raw QA set \(\mathcal{D}=\{(x,a,E)\}\) with question \(x\), answer \(a\), and supporting evidence \(E\);
  construction prompt \(\mathcal{P}_{\mathrm{gen}}\); annotator model \(g\)
\ENSURE
  Stepwise training set \(\mathcal{D}_{\mathrm{step}}\) with segmented answers and step-level supports;
  context maps \((\mathrm{ctx2id},\mathrm{id2ctx})\)
\STATE \(\widehat{\mathcal{D}}\gets \operatorname{Normalize}(\mathcal{D})\)
  \COMMENT{convert to a shared document-indexed schema}
\STATE \(\widehat{\mathcal{D}}\gets \operatorname{FilterUnsupported}(\widehat{\mathcal{D}})\)
  \COMMENT{remove examples without valid answer or support}
\STATE \((\mathrm{ctx2id},\mathrm{id2ctx})\gets \operatorname{BuildContextMaps}(\widehat{\mathcal{D}})\)
\STATE \(\mathcal{D}_{\mathrm{step}}\gets \emptyset\)
\FOR{each example \(e=(x,a,E)\in\widehat{\mathcal{D}}\)}
  \STATE \(B\gets \operatorname{GoldDocs}(E)\)
    \COMMENT{evidence documents with local DOC labels}
  \STATE \(r\gets g(\mathcal{P}_{\mathrm{gen}},B,x,a)\)
    \COMMENT{generate document-grounded answer segments}
  \STATE \(\{(y_s,d_s)\}_{s=0}^{T-1}\gets \operatorname{ParseDocSegments}(r)\)
    \COMMENT{\(s\): step index; \(T\): number of steps; \(y_s\): answer segment; \(d_s\): local DOC label}
  \STATE \(\{\mathcal{R}_s\}_{s=0}^{T-1}\gets \operatorname{AlignSupport}(\{d_s\}_{s=0}^{T-1},B,\mathrm{ctx2id},\mathrm{id2ctx})\)
  \STATE \(\mathbf{y}\gets [x,y_0,\langle m\rangle,y_1,\ldots,\langle m\rangle,y_{T-1}]\)
  \STATE \(\mathcal{D}_{\mathrm{step}}\gets \mathcal{D}_{\mathrm{step}}\cup\{(x,\mathbf{y},\{(y_s,\mathcal{R}_s)\}_{s=0}^{T-1},a)\}\)
\ENDFOR
\RETURN \(\mathcal{D}_{\mathrm{step}}, (\mathrm{ctx2id}, \mathrm{id2ctx})\)
\end{algorithmic}
\end{algorithm}

\begin{table*}[t]
\centering
\scriptsize
\renewcommand{\arraystretch}{1.08}
\setlength{\tabcolsep}{3pt}
\begin{tabular}{cp{0.16\textwidth}p{0.28\textwidth}p{0.43\textwidth}}
\toprule
\textbf{\(\langle m\rangle\) Tokens} & \textbf{Dataset / Answer} & \textbf{Question} & \textbf{Stepwise Target} \\
\midrule
0 &
NaturalQuestions / Tobey &
Who does Bo Burnham play in \textit{Rough Night}? &
Bo Burnham is listed as playing Tobey in \textit{Rough Night}. In conclusion, the answer is \(\boxed{\text{Tobey}}\). \\
\midrule
1 &
2WikiMQA / 20 July 2019 &
When did Mario Merz's wife die? &
Mario Merz was the husband of Marisa Merz. \(\langle m\rangle\)
Marisa Merz died on 20 July 2019. Therefore, Mario Merz's wife died on \(\boxed{\text{20 July 2019}}\). \\
\midrule
2 &
MuSiQue / Copenhagen &
Where did the spouse of the composer of \textit{Aladdin} die? &
The composer of \textit{Aladdin} is Carl Nielsen. \(\langle m\rangle\)
Carl Nielsen's spouse was Anne Marie Carl-Nielsen. \(\langle m\rangle\)
Anne Marie Carl-Nielsen died in Copenhagen. In conclusion, the answer is \(\boxed{\text{Copenhagen}}\). \\
\midrule
3 &
2WikiMQA / Pink Dream &
Which film has the director who died earlier, \textit{Pink Dream} or \textit{Bullet (1985 film)}? &
\textit{Pink Dream} was directed by Cai Chusheng. \(\langle m\rangle\)
\textit{Bullet (1985 film)} was directed by Bapu. \(\langle m\rangle\)
Cai Chusheng died on July 15, 1968. \(\langle m\rangle\)
Bapu died on August 31, 2014. Therefore, the film whose director died earlier is \(\boxed{\text{Pink Dream}}\). \\
\bottomrule
\end{tabular}
\caption{Representative processed training examples with zero to three inserted memory tokens. Each \(\langle m\rangle\) marks a refresh boundary between adjacent answer segments.}
\label{tab:training-data-examples}
\end{table*}

\subsection{Training Data Statistics}
\label{app:training_data_statistics}

Table~\ref{tab:training-data-statistics} reports the statistics of the processed training file used in our main experiments.
The final training set contains 32,319 examples.
The average number of refresh-supervised steps is 2.23, with a maximum of four steps.
NaturalQuestions contributes single-step examples, while 2WikiMQA and MuSiQue provide more diverse multi-step supervision.

\begin{table*}[t]
\centering
\small
\setlength{\tabcolsep}{3.5pt}
\begin{tabular}{lrrrrr}
\toprule
\textbf{Dataset} & \textbf{Examples} & \textbf{Average Steps} & \textbf{Median Steps} & \textbf{Maximum Steps} & \textbf{Average Documents} \\
\midrule
2WikiMQA & 12,567 & 2.89 & 2.0 & 4 & 2.89 \\
HotpotQA & 7,438 & 2.00 & 2.0 & 2 & 2.00 \\
MuSiQue & 5,424 & 2.59 & 3.0 & 4 & 2.59  \\
NaturalQuestions & 6,890 & 1.00 & 1.0 & 1 & 1.00 \\
\midrule
Total & 32,319 & 2.23 & 2.0 & 4 & 2.23 \\
\bottomrule
\end{tabular}
\caption{Statistics of the processed stepwise training data. Average Documents counts the average number of unique gold context documents used by each example. The Total row aggregates statistics over all examples rather than averaging dataset-level rows.}
\label{tab:training-data-statistics}
\end{table*}

\begin{table*}[t]
\centering
\small
\renewcommand{\arraystretch}{1}
\setlength{\tabcolsep}{5pt}
\begin{tabular}{lp{0.78\linewidth}}
\toprule
\textbf{Method} & \textbf{Inference setting} \\
\midrule
PReM & Chunk size 100. The 16× setting preserves top-20 chunks, and the 32× setting preserves top-10 chunks. \\
SnapKV & Per-layer KV budget 2,048 for 16× and 1,024 for 32×. The recent-window size is 8, with max-pooling and kernel size 7 for attention-score pooling. \\
StreamingLLM & Per-layer KV budget 2,048 for 16× and 1,024 for 32×. The method keeps 4 attention sink tokens and uses the remaining budget as the recent sliding window. \\
CAKE & Average per-layer KV budget 2,048 for 16× and 1,024 for 32×. CAKE adaptively allocates the cache budget across layers and keeps a recent-window size of 8 per layer. Other settings follow the official Qwen defaults, with \(\tau_1=1.5\), \(\tau_2=0.6\), and \(\gamma=200\). \\
LongLLMLingua & Text-compression target 2,048 tokens for 16× and 1,024 tokens for 32×. The compressor uses the same backbone as the evaluated model, and the dynamic compression ratio is set to 0.3. \\
LLMLingua-2-large & Text-compression target 2,048 tokens for 16× and 1,024 tokens for 32×. We use the released large version of LLMLingua-2 for text compression. \\
EXIT & Text-compression target 2,048 tokens for 16× and 1,024 tokens for 32×. We use the released Gemma-2B-based EXIT compressor with the query-conditioned ``Yes'' probability. \\
Activation Beacon & Released \texttt{beacon-qwen-2-7b-instruct} checkpoint with \texttt{beacon\_ratio} set to 16 or 32, together with the released adaptive minimum beacon setting, and the direct-output prompt. \\
ICAE & Released ICAE checkpoint on Mistral-7B-Instruct-v0.2, memory size 128, compression rates 16 and 32, and the direct-output prompt. \\
\bottomrule
\end{tabular}
\caption{Baseline-specific inference settings used for the main comparison. The 16× and 32× budgets correspond to 2K and 1K retained KV or context tokens under the 32K-token context setting.}
\label{tab:baseline-inference-settings}
\end{table*}

\section{Inference Settings}
\label{app:baseline-inference-settings}

\begin{figure}[h]
\centering
\includegraphics[width=0.45\textwidth]{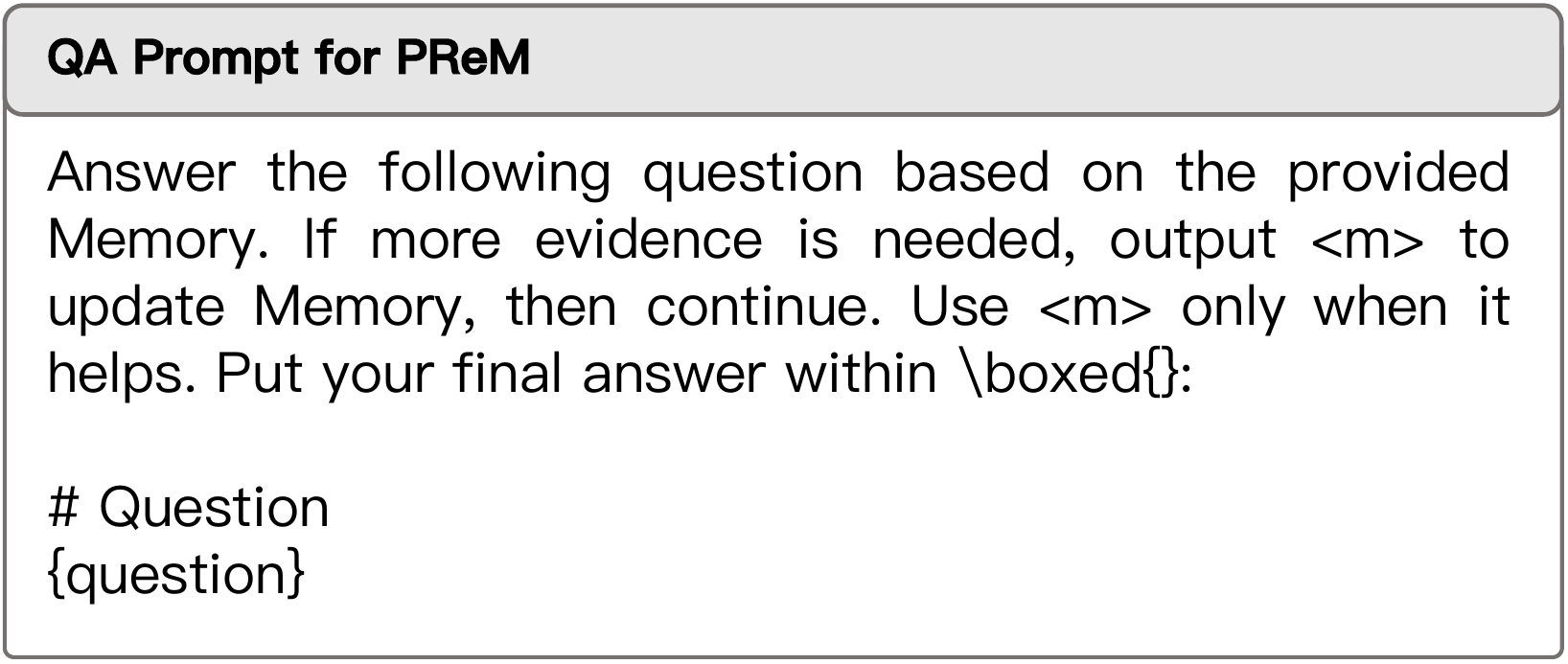}
\caption{Prompt template used for PReM inference.}
\label{fig:prem-inference-prompt}
\end{figure}

\begin{figure}[h]
\centering
\includegraphics[width=0.45\textwidth]{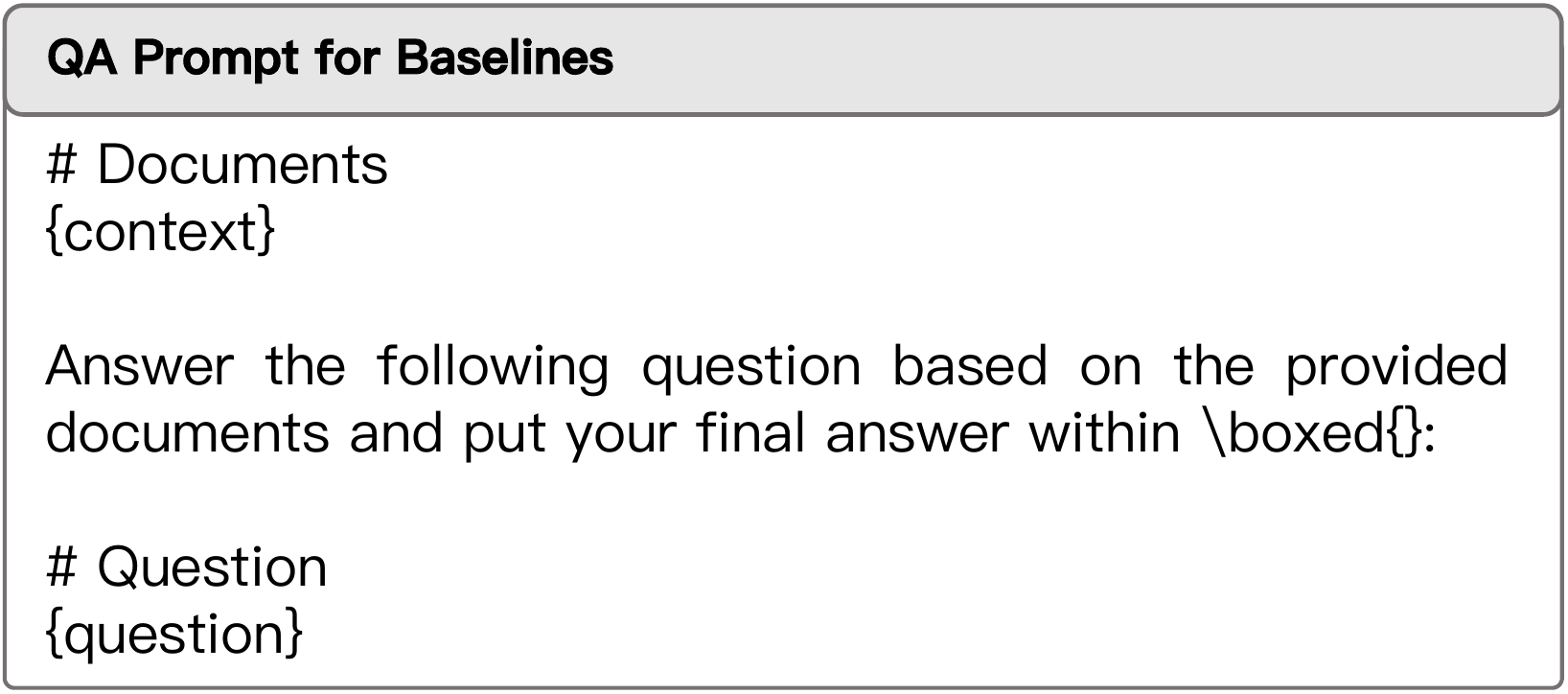}
\caption{Prompt template used for Qwen-matched prompting, KV-cache compression, and text-compression baselines.}
\label{fig:baseline-inference-prompts}
\end{figure}

\begin{figure}[h]
\centering
\includegraphics[width=0.45\textwidth]{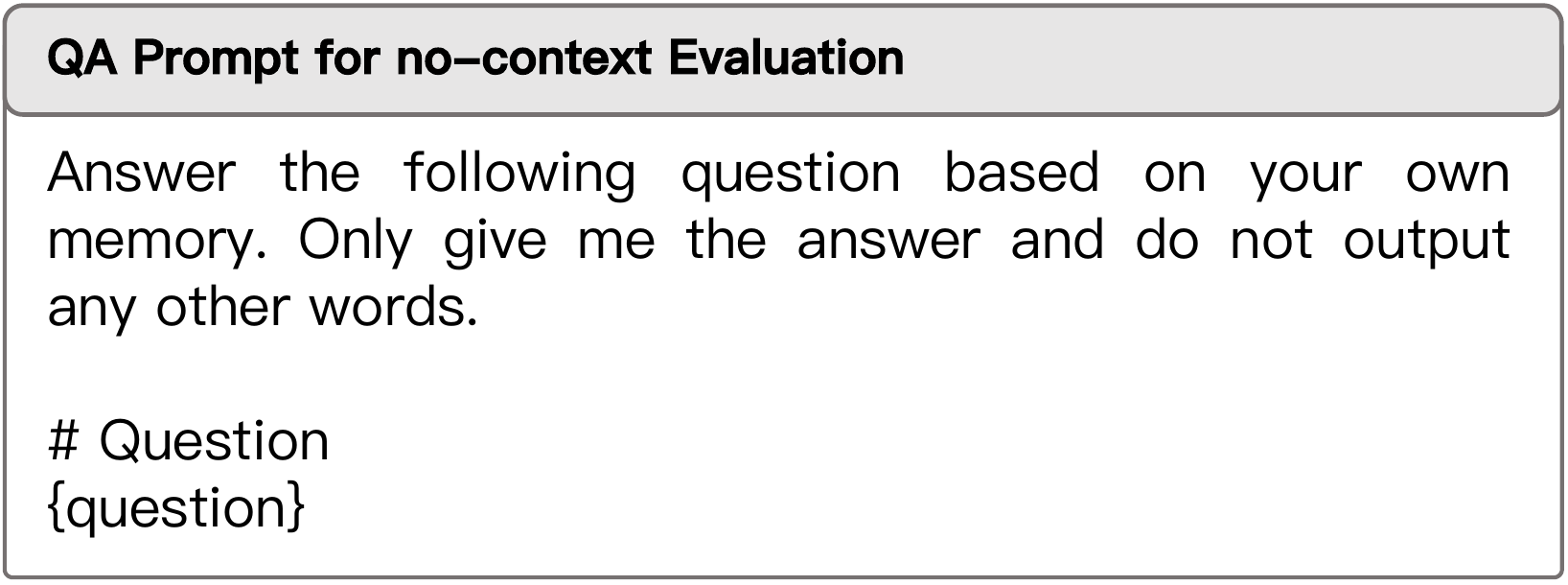}
\caption{Prompt template for no-context evaluation.}
\label{fig:baseline-no-context-prompt}
\end{figure}

\begin{figure}[h]
\centering
\includegraphics[width=0.45\textwidth]{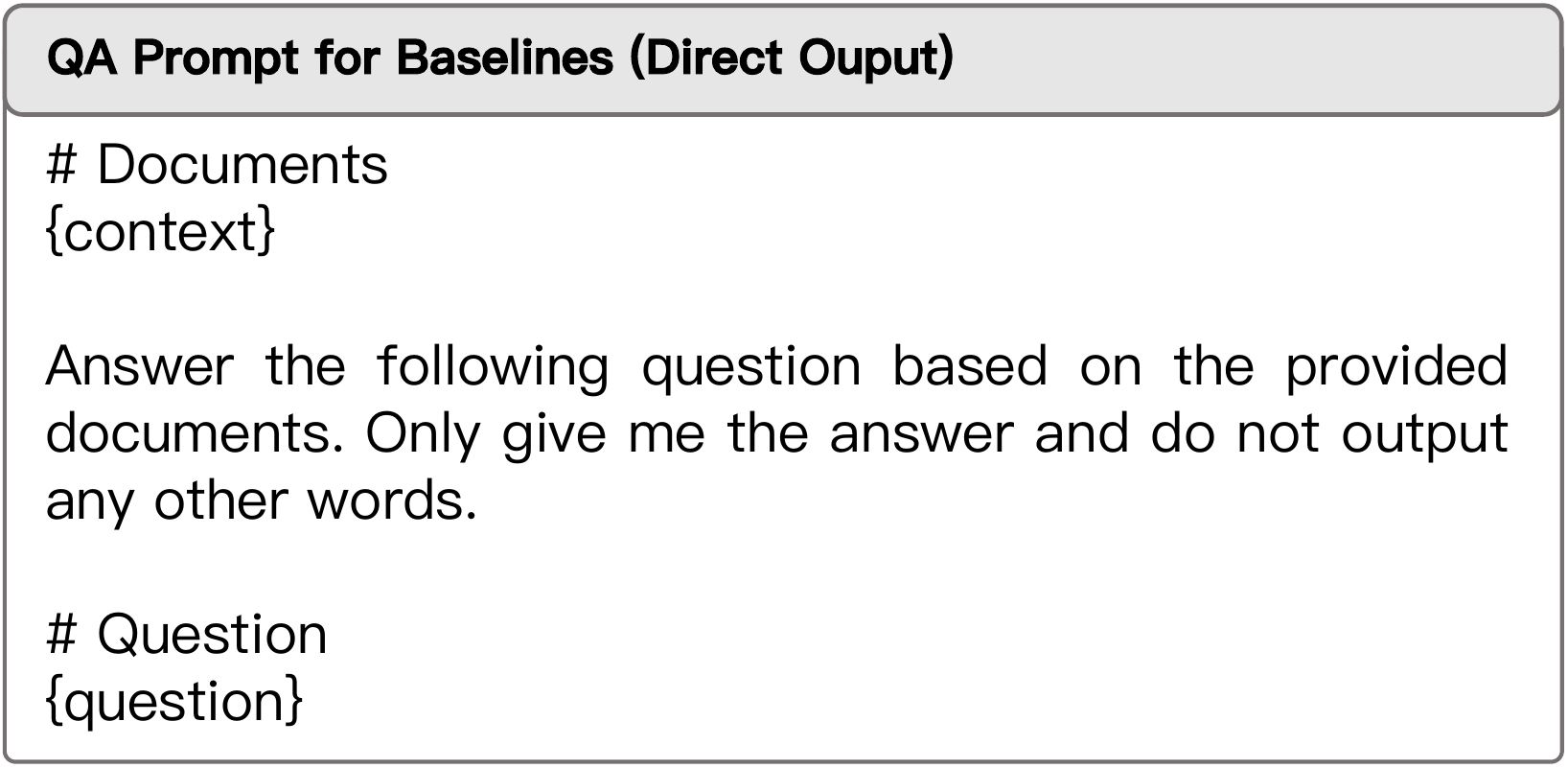}
\caption{Direct-output prompt template used for released soft context-compression baselines.}
\label{fig:baseline-direct-output-prompt}
\end{figure}

Table~\ref{tab:baseline-inference-settings} summarizes the inference settings.
All baselines are evaluated on the same 32K-token contexts constructed from gold documents and distractors, with random document-level placement of gold evidence, greedy decoding, and a maximum generation length of 512 tokens.
For consistency, context length is measured using the Qwen tokenizer for all methods.
Evaluation examples are randomly sampled from instances with near-zero no-context performance by Qwen2.5-7B-Instruct, where both EM and F1 are below 1 point, reducing reliance on parametric knowledge.
Qwen-matched baselines use the boxed-answer prompt for EM/F1 extraction.
For released soft context-compression baselines, namely Activation Beacon and ICAE, we use a direct-output prompt because their released checkpoints are intended for concise answer-style inference.
The prompt used by PReM is shown in Figure~\ref{fig:prem-inference-prompt}, the standard boxed-answer baseline prompt is shown in Figure~\ref{fig:baseline-inference-prompts}, the no-context evaluation prompt is shown in Figure~\ref{fig:baseline-no-context-prompt}, and the direct-output prompt for released soft context-compression baselines is shown in Figure~\ref{fig:baseline-direct-output-prompt}.

\section{Case Studies}
\label{app:qualitative-case-studies}

Table~\ref{tab:qualitative-case-studies} presents three representative cases from the 32× Qwen2.5-7B evaluation where PReM answers correctly but LongLLMLingua, LLMLingua-2-large, and EXIT fail.
These cases illustrate three typical failure modes of text-space compression: the compressed context may miss the document that directly contains the answer, keep only the first clue while dropping the document needed for the next reasoning step, or keep an early entity in a multi-hop chain while losing the later documents that connect it to the final answer.
This contrast indicates that dynamic memory refresh helps avoid these evidence-loss failures.

\begin{enumerate}
    \item \textbf{Missing answer-bearing document.}
    The compressor keeps loosely related fragments but drops the document that contains the final answer.
    \item \textbf{First-hop retention without target evidence.}
    The compressed context preserves the first-hop entity, but removes the second document needed to answer the question.
    \item \textbf{Broken long evidence chain.}
    The compressor keeps an early entity from a multi-hop question, but drops the later documents needed to connect that entity to the final answer.
\end{enumerate}

\begin{table*}[t]
\centering
\scriptsize
\setlength{\tabcolsep}{3pt}
\begin{tabular}{p{0.13\textwidth}p{0.25\textwidth}p{0.30\textwidth}p{0.24\textwidth}}
\toprule
Failure type & Question / answer & Compressed-context failure & Gold context evidence \\
\midrule
1. Missing answer-bearing document &
\textbf{HotpotQA}: Who directed the film that had ``The Distance'' in the soundtrack? \newline
Gold/PReM: Peter Chelsom &
All three text compressors remove the key answer string from the compressed context. The retained fragments are unrelated soundtrack or film snippets, and the baselines answer unknown or not stated. &
\textit{Evan and Jaron (album)} links ``The Distance'' to the \textit{Serendipity} soundtrack. \textit{Serendipity (film)} states that the film was directed by Peter Chelsom. \\
\midrule
2. First-hop retained, target evidence dropped &
\textbf{2WikiMQA}: What is the date of birth of the director of film \textit{A Single Shot}? \newline
Gold/PReM: March 23, 1969 &
The compressed contexts keep \textit{A Single Shot}, but drop the target biography \textit{David M. Rosenthal (director)} and the date ``March 23, 1969''. The baselines therefore return not provided or an empty answer. &
\textit{A Single Shot} identifies David M. Rosenthal as the director. \textit{David M. Rosenthal (director)} gives his birth date as March 23, 1969. \\
\midrule
3. Broken long evidence chain &
\textbf{MuSiQue}: What major conflict is the country between the country that hosted the tournament and the country where That Dam is from known for? \newline
Gold/PReM: one of the world's longest-running ongoing civil wars &
The compressed contexts keep the anchor \textit{That Dam}, but drop the bridging and final documents \textit{Geography of Myanmar} and \textit{Myanmar}. The answer phrase is absent; EXIT instead retains an unrelated World War II fragment. &
\textit{2020 AFC U-23 Championship qualification} identifies Thailand as host, \textit{That Dam} is in Laos, \textit{Geography of Myanmar} places Myanmar between Thailand and Laos, and \textit{Myanmar} states that the country has been involved in one of the world's longest-running ongoing civil wars. \\
\bottomrule
\end{tabular}
\caption{Evidence-level case studies illustrating three representative failure modes of text-space compression under 32× compression: missing the answer-bearing document, retaining only first-hop evidence, and breaking a longer evidence chain.}
\label{tab:qualitative-case-studies}
\end{table*}

Overall, these examples show that text-space compressors can discard crucial evidence before generation begins, especially when the answer depends on a small answer-bearing document or on a chain of multiple documents.
By contrast, PReM answers these cases correctly, suggesting that refreshed KV memory helps retain useful evidence even when the required information is sparse within the 32K context.

\ifdefined\mainpaperinput
\else

\end{document}
\fi


\end{document}